# AdaDepth: Unsupervised Content Congruent Adaptation for Depth Estimation

Jogendra Nath Kundu*   Phani Krishna Uppala*   Anuj Pahuja   R. Venkatesh Babu
Video Analytics Lab, Department of Computational and Data Sciences
Indian Institute of Science, Bangalore, India
jogendrak@iisc.ac.in, {krishnaphaniiitg, anujpahuja13}@gmail.com, venky@iisc.ac.in

## Abstract

*Supervised deep learning methods have shown promising results for the task of monocular depth estimation; but acquiring ground truth is costly, and prone to noise as well as inaccuracies. While synthetic datasets have been used to circumvent above problems, the resultant models do not generalize well to natural scenes due to the inherent domain shift. Recent adversarial approaches for domain adaption have performed well in mitigating the differences between the source and target domains. But these methods are mostly limited to a classification setup and do not scale well for fully-convolutional architectures. In this work, we propose AdaDepth - an unsupervised domain adaptation strategy for the pixel-wise regression task of monocular depth estimation. The proposed approach is devoid of above limitations through a) adversarial learning and b) explicit imposition of content consistency on the adapted target representation. Our unsupervised approach performs competitively with other established approaches on depth estimation tasks and achieves state-of-the-art results in a semi-supervised setting.*

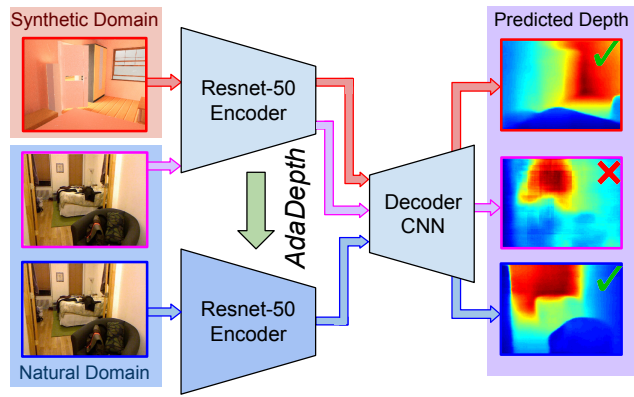

Figure 1. Illustration of the proposed domain adaptation method with input image domain discrepancy (red and blue background) followed by depth-map prediction. Color coded arrows represent corresponding RGB image and depth predictions for the synthetic-trained encoder (red and pink bordered) and for the adapted encoder (blue bordered); indicating that synthetic-trained model shows sub-optimal performance on natural images.

## 1. Introduction

Deep neural networks have brought a sudden sense of optimism for solving challenging computer vision tasks, especially in a data-hungry supervised setup. However, the generalizability of such models relies heavily on the availability of accurate annotations for massive amount of diverse training samples. To disentangle this dependency, researchers have started focusing towards the effectiveness of easily obtainable synthetic datasets in training deep neural models. For problem domains like semantic scene understanding, which face difficulty due to insufficient ground-truth for supervision, use of graphically rendered images has been a primary alternative. Even though synthetic images look visually appealing, deep models trained on them often perform sub-optimally when tested on real scenes, showing lack of generalization [19, 35]. From a probabilistic perspective, considering input samples for a network being drawn from a certain source distribution, the network can perform sufficiently well on test set only if the test data is also sampled from the same distribution. Hence, the general approach has been to transfer learned representations from synthetic to real datasets by fine-tuning the model on a mixed set of samples [42].

For depth estimation tasks, the ground-truth acquired using devices like Kinect or other depth sensors exhibits noisy artifacts [40] and hence severely limits the performance of a supervised depth prediction network. In the widely used NYU Depth Dataset [34], such cases are addressed by manually inpainting the depth values in the distorted regions. But the dataset has only a handful of such crafted samples, mainly because the process is laborious and prone to pixel-level annotation errors. These shortcomings show the need for a framework that is minimally dependent on scarce clean ground truth data. *AdaDepth* addresses this need by adapting representations learned from graphically rendered synthetic image and depth pairs to real natural scenes.

---
*Equal contribution

Monocular depth estimation is an ill-posed problem; yet it has many applications in graphics [21], computational photography [2] and robotics [26, 41]. To overcome the lack of multi-view information, depth prediction models need to exploit global semantic information to regress accurate pixel-level depth. It is observed that an end-to-end Fully Convolutional Network (FCN) [25] can extract useful objectness features for efficient depth prediction without explicit enforcement. Such objectness information is exhibited by both synthetic and natural scenes as synthetic scenes also adhere to the natural distribution of relative object placement.

Previous works on domain adaptation techniques either attempt to learn an extra mapping layer to reduce domain representation gap [33] or learn domain invariant representations by simultaneously adapting for both source and target domains [44]. In contrast to classification-based approaches, there are very few works focusing on spatially structured prediction tasks [17]. Zhang *et al*. [50] show the inefficiency of classification-based approaches on such tasks, mostly because of the higher dimensional feature space. To the best of our knowledge, we are the first to explore unsupervised adversarial domain adaptation for a spatially structured regression task of depth estimation. In general, *Mode collapse* [37] is a common phenomenon observed during adversarial training in absence of paired supervision. Because of the complex embedded representation of FCN, preservation of spatial input structure in an unsupervised adaptation process becomes challenging during adversarial learning. Considering no access to target depth-maps, we address this challenge using the proposed *content congruent regularization* methods that preserve the input structural content during adaptation. The proposed adaptation paradigm results in improved depth-map estimation when tested on the target natural scenes.

Our contributions in this paper are as follows:

- We propose an unsupervised adversarial adaptation setup *AdaDepth*, that works on the high-dimensional structured encoder representation in contrast to adaptation at task-specific output layer.
- We address the problem of *mode collapse* by enforcing content consistency on the adapted representation using a novel feature reconstruction regularization framework.
- We demonstrate *AdaDepth's* effectiveness on the task of monocular depth estimation by empirically evaluating on NYU Depth and KITTI datasets. With minimal supervision, we also show state-of-the-art performance on depth estimation for natural target scenes.

## 2. Related work

**Supervised Monocular Depth Estimation** There is a cluster of previous works on the use of hand-crafted features and probabilistic models to address the problem of depth estimation from single image. Liu *et al*. [28] use predicted labels from semantic segmentation to explicitly use the objectness cues for the depth estimation task. Ladicky *et al*. [24] instead carry out a joint prediction of pixel-level semantic class and depth. Recent spurt in deep learning based methods has motivated researchers to use rich CNN features for this task. Eigen *et al*. [6] were the first to use CNNs for depth regression by integrating coarse and fine scale features using a two-scale architecture. They also combined the prediction of surface normals and semantic labels with a deeper VGG inspired architecture with three-scale refinement [5]. To further improve the prediction quality, hierarchical graphical models like CRF have been combined with the CNN based super-pixel depth estimation [27]. For continuous depth prediction, Liu *et al*. [29] use deep convolutional neural fields to learn the end-to-end unary and pairwise potentials of CRF to facilitate the training process. Laina *et al*. [25] proposed a ResNet [16] based encoder-decoder architecture with improved depth prediction results.

**Unsupervised/Semi-supervised Depth Estimation** Another line of related work on depth estimation focuses on unsupervised/semi-supervised approaches using geometry-based cues. Garg *et al*. [10] proposed an encoder-decoder architecture to predict depth maps from stereo pair images using an image alignment loss. Extending this, Godard *et al*. [13] proposed to minimize the left-right consistency of estimated disparities in stereo image pair for the unsupervised depth prediction task. On the other hand, Yevhen *et al*. [23] follow a semi-supervised approach using sparse ground-truth depth-map along with the image alignment loss in a stereo matching setup. Zhou *et al*. [52] used video sequences for depth prediction with view synthesis as a supervisory signal.

**Transfer learning using Synthetic Scenes** Lately, graphically rendered datasets are being used for various computer vision tasks such as pose prediction of human and objects [42, 47], optical flow prediction [4] and semantic segmentation [35]. Zhang *et al*. [51] proposed a large-scale physically-based rendering dataset for indoor scenes to bridge the gap between the real and synthetic scene with improved lighting setup. But training deep CNN models on such diverse synthetic images does not generalize directly for natural RGB scenes.

**Domain adaptation** Csurka [3] published a comprehensive survey on domain adaptation techniques for visual applications. Our work falls in the subarea of DeepDA (Deep Domain Adaptation) architectures. Several such architec-tures incorporate a classification loss and a discrepancy loss [12, 46, 31, 43], with Maximum Mean Discrepancy (MMD) [15] being the commonly used discrepancy loss. Long *et al.* [31] use MMD for the layers embedded in a kernel Hilbert space to effectively learn the higher order statistics between the source and target distribution. Sun and Saenko [43] proposed a deep correlation alignment algorithm (CORAL) which matches the mean and covariance of the two distributions at the final feature level to align their second-order statistics for adaptation.

Another line of work uses adversarial loss in conjunction with classification loss, with an objective to diminish domain confusion [44, 8, 9, 45]. As opposed to prior works that usually use a fully-connected layer at the end for class adaptation, we employ a DeepDA architecture for a more challenging pixel-wise regression task of depth estimation. Our proposed method uses the concept of Generative Adversarial Networks (GANs) [14] to address the domain discrepancy at an intermediate feature level. In GAN framework, the objective of generator is to produce data which can fool the discriminator, whereas the discriminator improves itself by discriminating the generated samples from the given target distribution. Following this, Isola *et al.* [18] proposed *pix2pix*, that uses a conditional discriminator to enforce consistency in generated image for a given abstract representation. Without such conditioning, the generator can produce random samples that are inconsistent with the given input representation, while minimizing the adversarial loss. As an extension, Zhu *et al.* [53] introduced *CycleGAN*, a cycle consistency framework to enforce consistency of input representation at the generator output for unpaired image-to-image translation task.

## 3. Approach

Consider synthetic images $x_s \in X_s$ and the corresponding depth maps $y_s \in Y_s$ as samples from a source distribution, $p_s(x, y)$. Similarly, the real images $x_t \in X_t$ are considered to be drawn from a target distribution $p_t(x, y)$, where $p_s \neq p_t$. Under the assumption of unsupervised adaptation, we do not have access to the real depth samples $y_t \in Y_t$.

Considering a deep CNN model as a transfer function from an input image to the corresponding depth, the base model can be divided into two transformations: $M_s$, that transforms an image to latent representation, and $T_s$, that transforms latent representation to the final depth prediction. The base CNN model is first trained with full supervision from the available synthetic image-depth pairs i.e. $\bar{y}_s = T_s(M_s(x_s))$. A separate depth prediction model for the real images drawn from target distribution can be written as $\bar{y}_t = T_t(M_t(x_t))$. Due to *domain shift*, direct inference on target samples $x_t$ through the network trained on $X_s$ results in conflicting latent representation and predictions, i.e.

$M_s(x_t) \neq M_t(x_t)$ and $T_s(M_s(x_t)) \neq T_t(M_t(x_t))$. For effective domain adaptation, ideally both $M_s$ and $T_s$ have to be adapted to get better performance for the target samples. Considering that $X_s$ and $X_t$ only exhibit perceptual differences caused by the graphical rendering process, both domains have many similarities in terms of objectness information and relative object placement. Therefore, we only adapt $M_t$ for the target distribution $p_t(x)$. To generalize the learned features for the new domain, we plan to match the latent distributions of $M_s(X_s)$ and $M_t(X_t)$ so that the subsequent transformation $T_s$ can be used independent of the domain as $T_s = T_t = T$.

We start the adaptation process by initializing $M_t$ and $T_t$ with the supervisely trained weights from $M_s$ and $T_s$ respectively. To adapt the parameters of $M_t$ for the target samples $x_t$, we introduce two different discriminators $D_F$ and $D_Y$. The objective of $D_F$ is to discriminate between the source and target latent representations $M_s(x_s)$ and $M_t(x_t)$, whereas the objective of $D_Y$ is to discriminate between $Y_s$ and $T(M_t(X_t))$. Assuming similar depth map distribution for both synthetic and real scenes ($p(Y_s = y_s) \approx p(Y_t = y_t)$), inferences through the corresponding transformation functions $T(M_s(x_s))$ and $T(M_t(x_t))$ are directed towards the same output density function.

We use a ResNet-50 [16] based encoder-decoder architecture [25] for demonstrating our approach. Existing literature [49] reveals that in hierarchical deep networks, the lower layers learn generic features related to the given data distribution whereas the consequent layers learn more task specific features. This implies that the transferability of learned features for different data distributions (source and target) decreases as we move from lower to higher layers with an increase in domain discrimination capability. We experimentally evaluated this by varying the number of shared layers between $M_s$ and $M_t$, starting from the initial layers to the final layers. From Figure 3, it is clear that towards higher layers of $M_s$, features are more discriminable for synthetic versus natural input distribution. Therefore, we deduce that adaptation using only *Res-5* blocks of $M_t$ (*Res-5a*, *Res-5b* and *Res-5c*) and fixed shared parameters of other layers (Figure 2) is optimal for adversarial adaptation as it requires minimal number of parameters to update.

In rest of this section, we describe the adversarial objectives along with the proposed content consistent loss formulations to update the parameters of $M_t$ for depth estimation.

### 3.1. Adversarial Objectives

We define an adversarial objective $L_{advD}$ at the prediction level for $D_Y$ and an adversarial objective $L_{advF}$ at the latent space feature level for $D_F$. They can be defined as:

$$\mathcal{L}_{advD} = \mathbb{E}_{y_s \sim Y_s}[\log D_Y(y_s)] \\ + \mathbb{E}_{x_t \sim X_t}[\log(1 - (D_Y(T(M_t(x_t)))))] \quad (1)$$

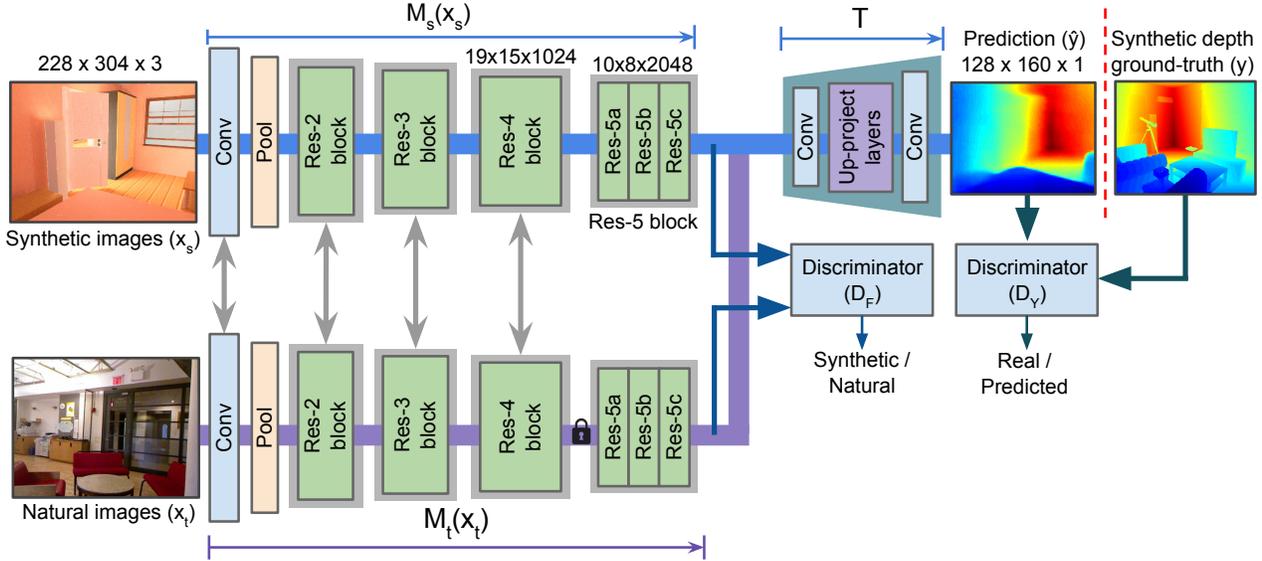

Figure 2. *AdaDepth*: Our deep residual encoder-decoder base architecture with adversarial setup illustrating different transformation functions as described in Section 3. The source (synthetic) and target (real) branch are specified by blue and purple channel respectively. The double-headed arrows between res-blocks indicate parameter sharing. Note that during adaptation of the synthetic-trained $T(M_t(x_t))$, only the layers in purple branch are updated (i.e. *Res-5* block) until the location of *lock icon*.

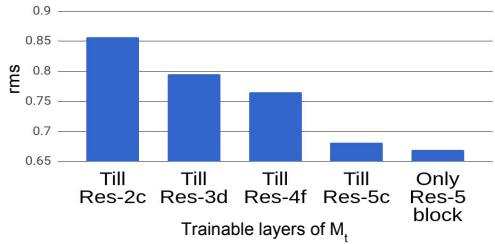

Figure 3. Effect of various weight sharing strategies on adversarial adaptation process with domain consistency regularization (Section 3.2.1).

$$\mathcal{L}_{advF} = \mathbb{E}_{x_s \sim X_s}[\log D_F(M_s(x_s))] \\ + \mathbb{E}_{x_t \sim X_t}[\log(1 - (D_F(M_t(x_t))))] \quad (2)$$

$M_t$ parameters are updated to minimize both the adversarial losses, whereas the discriminators $D_Y$ and $D_F$ are updated to maximize the respective objective functions. The final objective to update the parameters of $M_t$, $D_Y$ and $D_F$ can be expressed as $\min_{M_t} \max_{D_Y} \mathcal{L}_{advD}$ and $\min_{M_t} \max_{D_F} \mathcal{L}_{advF}$.

### 3.2. Content Congruency

In practice, a deep CNN exhibits complex output and latent feature distribution with multiple modes. Relying only on adversarial objective for parameter update leads to *mode collapse*. Theoretically, adversarial objective should work for a stochastic transfer function. However, since we do not use any randomness in our depth prediction model, it is highly susceptible to this problem. At times, the output prediction becomes inconsistent with the corresponding input image even at optimum adversarial objective. To tackle this, we enforce content congruent regularization methods as discussed below.

#### 3.2.1 Domain Consistency Regularization (DCR)

Since we start the adversarial learning after training on synthetic images, the resultant adaptation via adversarial objective should not distort the rich learned representations from the source domain. It is then reasonable to assume that $M_s$ and $M_t$ differ by a small perturbation. We do so by enforcing a constraint on the learned representation while adapting the parameters for the new target domain. As per the proposed constraint, the latent representation for the samples from the target domain $M_t(x_t)$ must be regularized during the adaptation process with respect to $M_s(x_t)$ and can be represented as:

$$\mathcal{L}_{domain} = \mathbb{E}_{x_t \sim X_t}[\|M_s(x_t) - M_t(x_t)\|_1] \quad (3)$$

#### 3.2.2 Residual Transfer Framework (RTF)

Considering the adaptation process from $M_s$ to $M_t$ as a feature perturbation, Long *et al*. [32] proposed a residual transfer network to model $M_t$ as $M_s + \Delta M$. On similar lines, we implement an additional skip multi-layer CNN block with additive feature fusion to model $\Delta M$ such that $M_t = M_s + \Delta M$ (Figure 4a). To maintain content consistency, $\Delta M$ is constrained to be of low value so as to avoid distortion of the base $M_s$ activations. Also note that in this framework, the only trainable parameters for the adaptation process are $\Theta_{\Delta M}$, i.e. the parameters of the residual branch in Figure 4a. Considering $L_t(x_t)$ as the output feature activation after *Res-4* block, the regularization term can be written as:

$$\mathcal{L}_{res} = \mathbb{E}_{x_t \sim X_t}[\|\Delta M(L_t(x_t))\|_2] \quad (4)$$

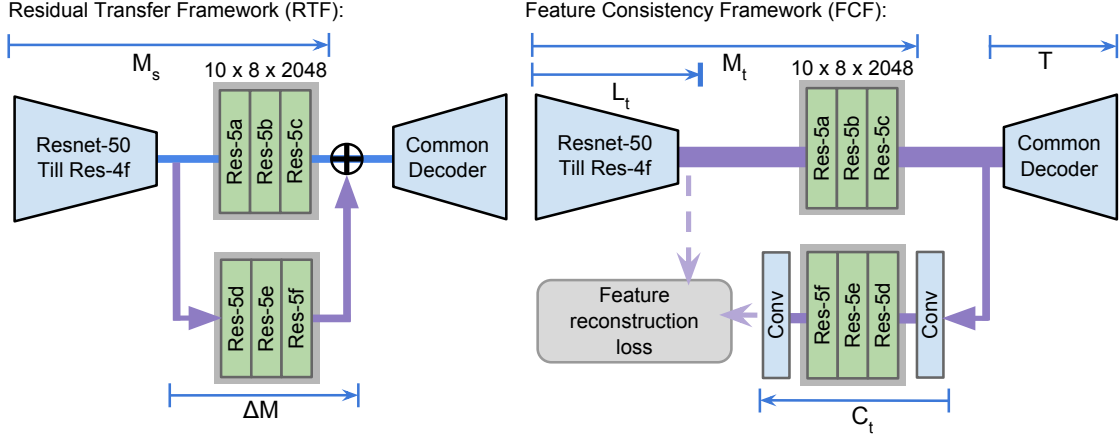

Figure 4. Overview of the proposed a) Residual Transfer Framework (left) and b) Feature Consistency Framework (right) described in Section 3.2.2 and 3.2.3 respectively. The branches with purple background indicate trainable layers during the adaptation process.

### 3.2.3 Feature Consistency Framework (FCF)

As a new approach to preserve spatial structure and content correspondence between the input image and the predicted depth map, we propose to enforce content consistency using a self feature reconstruction loss. We formulate feature consistency of *Res-5* block to regularize the adversarial adaptation process which can respect the corresponding depth prediction. We define $C_t$ as a parameterized feature reconstruction function (a multi-layer CNN) to reconstruct the *Res-4f* features while updating the trainable parameters of $M_t$ using adversarial discriminator loss. Mathematically, the regularization term is represented as:

$$\mathcal{L}_{feature} = \mathbb{E}_{x_t \sim X_t}[\|L_t(x_t) - C_t(M_t(x_t))\|_1] \quad (5)$$

## 3.3. Full objective

The final loss function while training $M_t$ is formulated as

$$\mathcal{L}_{final} = \mathcal{L}_{advD} + \mathcal{L}_{advF} + \lambda \mathcal{L}_{content} \quad (6)$$

where $\lambda$ is the weighting factor for the content regularization term relative to the adversarial losses, with $\mathcal{L}_{content}$ being one of the regularization methods (i.e. $\mathcal{L}_{domain}$, $\mathcal{L}_{res}$ or $\mathcal{L}_{feature}$). A lower $\lambda$ value increases the probability of *mode collapse*, whereas a higher $\lambda$ value enforces a limit to the adaptation process.

For Residual Transfer Framework, the search for appropriate hyperparameter $\lambda$ is even more difficult because of the uninitialized parameters introduced by $\Delta M$. Whereas for Feature Consistency Framework, $C_t$ is initialized with parameters trained to reconstruct $L_t(x_t)$ which significantly stabilizes the adversarial learning process. Algorithm 1 explains the adversarial learning strategy with the proposed Feature Consistency Framework.

We refer to the regularization frameworks mentioned in Section 3.2.1, Section 3.2.2 and Section 3.2.3 as DCR, RTF and FCF respectively for the rest of the paper.

/*Initialization of parameters */
$\Theta_{M_t}$: parameters of pretrained source encoder $M_s$
$\Theta_{C_t}$: parameters of fully trained $C_t$ branch by minimizing $\mathcal{L}_{feature}$, where $M_t = M_s$
$\Theta_{D_F}$: Randomly initialized weights
**for** $k$ *iterations* **do**
    **for** $m$ *steps* **do**
        $x_t$: minibatch sample of target images
        $x_s$: minibatch sample of source images
        /* Update parameters for $D_F$, $D_Y$ and $C_t$ */
        $\Theta^*_{D_F} := \underset{\Theta_{D_F}}{\operatorname{argmax}} \mathcal{L}_{advF}$
        $\Theta^*_{D_Y} := \underset{\Theta_{D_Y}}{\operatorname{argmax}} \mathcal{L}_{advD}$
        $\Theta^*_{C_t} := \underset{\Theta_{C_t}}{\operatorname{argmin}} \mathcal{L}_{feature}$
    **end**
    $x_t$: minibatch sample of target images
    $x_s$: minibatch sample of source images
    /* update trainable parameters of $M_t$ i.e. $\Theta_{M_t}$ */
    $\Theta^*_{M_t} := \underset{\Theta_{M_t}}{\operatorname{argmin}} (\mathcal{L}_{advF} + \mathcal{L}_{advD} + \lambda \mathcal{L}_{feature})$
**end**

**Algorithm 1:** Adversarial adaptation training algorithm for the proposed Feature Consistency Framework (FCF). The optimization steps are implemented using stochastic gradient updates of each minibatch.

| Method | Error metrics ↓ | | | Accuracy metrics ↑ | | |
|---|---|---|---|---|---|---|
| | rel | rms | $\log_{10}$ | $\delta_1$ | $\delta_2$ | $\delta_3$ |
| Baseline | 0.305 | 1.094 | 0.114 | 0.540 | 0.827 | 0.937 |
| Ours-DCR | 0.146 | 0.669 | 0.062 | 0.766 | 0.932 | 0.979 |
| Ours-RTF | 0.141 | 0.658 | 0.059 | 0.793 | 0.942 | 0.980 |
| Ours-FCF | **0.136** | **0.603** | **0.057** | **0.805** | **0.948** | **0.982** |
| FCF w/o $D_Y$ | 0.145 | 0.662 | 0.061 | 0.771 | 0.936 | 0.979 |

Table 1. *AdaDepth-U* results using different content consistency frameworks on NYU Depth [34] Test Set. For accuracy metrics, $\delta_i$ denotes $\delta_i < 1.25^i$ and higher is better.

| Method | Error metrics ↓ | | | Accuracy metrics ↑ | | |
|---|---|---|---|---|---|---|
| | rel | rms | log(rms) | $\delta_1$ | $\delta_2$ | $\delta_3$ |
| Baseline | 0.330 | 8.245 | 0.382 | 0.545 | 0.801 | 0.905 |
| Ours-DCR | 0.302 | 8.095 | 0.359 | 0.582 | 0.821 | 0.926 |
| Ours-RTF | 0.296 | 7.832 | 0.332 | 0.593 | 0.837 | 0.939 |
| Ours-FCF | **0.214** | **7.157** | **0.295** | **0.665** | **0.882** | **0.950** |

Table 2. *AdaDepth-U* results using different content consistency frameworks on Eigen Test Split of KITTI. [6] For accuracy metrics, $\delta_i$ denotes $\delta_i < 1.25^i$ and higher is better.

## 4. Experiments

In this section, we describe our implementation details and experiments on NYU Depth v2 [34] and KITTI [11] Datasets. We hereafter refer to our unsupervised and semi-supervised domain adaptation approaches as *AdaDepth-U* and *AdaDepth-S* respectively.

### 4.1. Network Architecture

For our base depth prediction network, we follow the architecture used by Laina *et al*. [25]. The network uses ResNet-50 [16] as the base encoder model followed by up-projection layers as shown in Figure 2. Similar to [25], we use BerHu (reverse Huber) loss to train the base network on synthetic images.

The extra convolutional branch $C_t$ and $\Delta M$ (Figure 4), used in feature reconstruction (FCF) and residual adaptation framework (RTF) respectively, constitutes residual blocks with batch-normalization layers similar to *Res-5* block. For the base network architecture, the output of $M_s(x_s)$ transformation is of spatial size 8×10, with 2048 activation channels. In contrast to fully-connected feature [45], we use spatial feature block (convolutional) as the latent representation during unsupervised adaptation. Hence, we implement $D_F$ as a standard convolutional discriminator architecture. For discriminator network $D_Y$, we follow Patch-GAN's [18] convolutional architecture with an input receptive field of size 80×80.

### 4.2. NYU Depth

**Dataset** NYU Depth v2 indoor scene dataset contains raw and clean RGB-D samples. The raw dataset consists of 464 scenes with a [249, 215] train-test split. The clean dataset comprises of 1449 RGB-D samples, where the depth maps are inpainted and aligned with RGB images. We use the commonly used test set of 654 images from these 1449 samples for final evaluation. Note that we do not use ground truth depth samples from the NYU Depth dataset for *AdaDepth-U*. For *AdaDepth-S*, we use 795 ground truth samples (6.5%) from the 1449 clean pairs. Both raw and clean samples have a spatial resolution of 480×640.

**Pre-Training** For pre-training our base network, we use 100,000 random samples ([80, 20] train-val split) out of 568,793 synthetic RGB-D pairs from the Physically-Based Rendering Dataset proposed by Zhang *et al*. [51]. Following [25], the input images of size 480×640 are first down-sampled by a factor of 2, and then center-cropped to size 228×304. Final prediction depth map is of spatial resolution 128×160.

**Evaluation** For comparison with ground truth, predictions up-sampled to the original size using bi-linear interpolation. We evaluate our final results by computing standard error and accuracy metrics used by [6, 25].

### 4.3. KITTI

**Dataset** KITTI dataset consists of more than 40,000 stereo pairs along with corresponding LIDAR data. We use the split proposed by [6] that contains 22,600 images for training and 697 images for testing. Note that we do not use any ground truth depth from the KITTI dataset for *AdaDepth-U*. For *AdaDepth-S*, we use 1000 random ground truth samples (4.4%) from the 22,600 images. All images have a spatial resolution of 375×1242.

**Pre-Training** For pre-training the base network, we use 21,260 synthetic RGB-D pairs provided in the Virtual KITTI Dataset [7]. We perform data augmentation on-the-fly similar to [13] during training. The input images of size 375×1242 are down-sampled to 256×512 before passing to the network. Final prediction depth map is of spatial resolution 128×256.

**Evaluation** In line with [13], we convert LIDAR data to depth images for comparison. We evaluate our final results by computing standard error and accuracy metrics used by [13, 52], with errors only being computed for depths less than 80 meters. We also evaluate our results with a cap of 50 meters for a fair comparison with [10].

### 4.4. Training Details

**Base Network** The base prediction model is trained from scratch for pre-training using TensorFlow [1]. During training, we use a mini-batch size of 10 and optimize with Adam [22]. We start with a high learning rate of 0.01, which is gradually reduced by a factor of 10 depending on the validation set performance.

| Method | Ground-truth Supervision | Error metrics ↓ | | | Accuracy metrics ↑ | | |
|---|---|---|---|---|---|---|---|
| | | rel | rms | $\log_{10}$ | $\delta < 1.25$ | $\delta < 1.25^2$ | $\delta < 1.25^3$ |
| Li et al. [27] | Yes | 0.232 | 0.821 | 0.094 | 0.621 | 0.886 | 0.968 |
| Liu et al. [30] | Yes | 0.230 | 0.824 | 0.095 | 0.614 | 0.883 | 0.971 |
| Wang et al. [48] | Yes | 0.220 | 0.745 | 0.094 | 0.605 | 0.890 | 0.970 |
| Eigen et al. [6] | Yes | 0.215 | 0.907 | - | 0.611 | 0.887 | 0.971 |
| Roy and Todorovic [36] | Yes | 0.187 | 0.744 | 0.078 | - | - | - |
| Eigen and Fergus [5] | Yes | 0.158 | 0.641 | - | 0.769 | 0.950 | 0.988 |
| Laina et al. [25] | Yes | 0.127 | 0.573 | 0.055 | 0.811 | 0.953 | 0.988 |
| DAN [31] | No | 0.281 | 0.859 | 0.095 | 0.583 | 0.848 | 0.946 |
| **Ours** AdaDepth-U(FCF) | No | 0.136 | 0.603 | 0.057 | 0.805 | 0.948 | 0.982 |
| **Ours** AdaDepth-S | Semi | **0.114** | **0.506** | **0.046** | **0.856** | **0.966** | **0.991** |

Table 3. Results on NYU Depth v2 Test Dataset. Baseline numbers have been taken from [25]. *AdaDepth-U* performs competitively with other methods while *AdaDepth-S* outperforms all of them. Note that all other methods use full ground truth supervision.

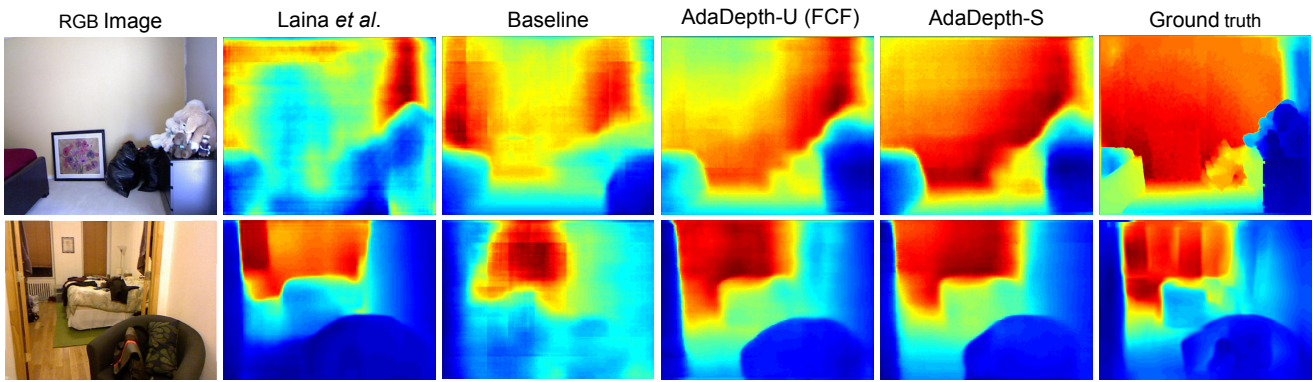

Figure 5. Qualitative comparison of *AdaDepth-U* and *AdaDepth-S* with Laina *et al*. [25]. Please refer to the supplementary material for additional results.

**Adaptation Network** During adaptation stage we use Momentum optimizer for updating the discriminator and generator parameters with a $\lambda$ value of 10. As mentioned in Algorithm 1, the parameters of $C_t$ are updated first to reconstruct the convolutional feature map of the penultimate *Res-4f* block before the adaptation process. Later, $C_t$ is updated along with $D_F$ and $D_y$ during the adversarial training to adapt $M_t$ for the new target domain. We also replace the adversarial binary-cross entropy formulation with least square loss in the adversarial objective, which further stabilizes the optimization process with better adaptation results.

The training of *AdaDepth-S* starts from the initialization of *AdaDepth-U* along with a very small set of target labeled data (Sections 4.2, 4.3). To avoid over-fitting, alternate batches of labeled (with ground-truth depth map) and unlabeled target samples are shown. For labeled batch iteration, we modify the final objective (Eq. 6) by replacing $L_{content}$ by BerHu loss computed between the predicted and ground-truth depth-map.

## 5. Results

In this section, we present a thorough evaluation of our proposed content consistency losses along with the adversarial objective functions as defined in Section 3.2 with a baseline approach. We also present comparative results of *AdaDepth-U* and *AdaDepth-S* with other depth prediction networks on NYU Depth V2 and KITTI datasets. Due to differences in scales between data domains, we scale our final predictions with a scalar $s = median(D_{gt})/median(D_{pred})$ for final evaluation, similar to Zhou *et al*. [52].

**Evaluation of content consistency methods** In Tables 1 and 2, we compare various design choices for our adversarial adaptation architecture by evaluating performance metrics using each of the regularization methods described in Section 3.2. As a baseline, we report the results on target (real) samples with direct inference on the network trained on source (synthetic) images. The metrics clearly demonstrate the advantage of adversarial domain adaptation with respect to baseline. Evidently, Feature Consistency Framework shows better performance as compared to other two techniques for unsupervised adaptation of both NYUD and KITTI natural datasets. During *mode collapse*, *Res-5* block learns a (non-invertible) many-to-one function and hence loses content information. The effectiveness of FCF over other two techniques can be attributed to explicit content preservation by learning the inverse function $C_t$ which makes it learn a one-to-one mapping during the unpaired adaptation process. We also do an ablation study without $D_Y$ (Table 1). It is evident from the experiment that $D_Y$ helps to preserve the continuous valued *depth-like* structure (ground-truth synthetic depth distribution) in the final prediction. Hence, $D_F$ along with $D_Y$ helps to bridge the domain discrepancy underlying both marginal ($P(M(x_s))$) and conditional distribution ($P(\hat{y}|M(x_s))$), which is crucial for domain adaptation [32].

| Method | Supervision | | Error metrics ↓ | | | | Accuracy metrics ↑ | | |
|---|---|---|---|---|---|---|---|---|---|
| | Depth | Pose | rel | sq. rel | rms | rms(log) | $\delta < 1.25$ | $\delta < 1.25^2$ | $\delta < 1.25^3$ |
| Eigen et al. [6] | Yes | No | 0.203 | 1.548 | 6.307 | 0.282 | 0.702 | 0.890 | 0.958 |
| Godard et al. [13] | No | Yes | **0.148** | 1.344 | 5.927 | 0.247 | **0.803** | **0.922** | 0.964 |
| Zhou et al. [52] | No | No | 0.208 | 1.768 | 6.856 | 0.283 | 0.678 | 0.885 | 0.957 |
| **Ours** AdaDepth-U(FCF) | No | No | 0.214 | 1.932 | 7.157 | 0.295 | 0.665 | 0.882 | 0.950 |
| **Ours** AdaDepth-S | Semi | No | 0.167 | **1.257** | **5.578** | **0.237** | 0.771 | **0.922** | **0.971** |
| Garg et al. [10] cap 50m | No | Yes | 0.169 | 1.080 | 5.104 | 0.273 | 0.740 | 0.904 | 0.962 |
| **Ours** AdaDepth-U cap 50m | No | No | 0.203 | 1.734 | 6.251 | 0.284 | 0.687 | 0.899 | 0.958 |
| **Ours** AdaDepth-S cap 50m | Semi | No | **0.162** | **1.041** | **4.344** | **0.225** | **0.784** | **0.930** | **0.974** |

Table 4. Results on KITTI Dataset using the Eigen test split [6]. Baseline numbers have been taken from [52]. With the exception of [52], all methods use either depth or pose ground truth supervision. *AdaDepth-U* shows comparable metrics to existing methods while *AdaDepth-S* outperforms existing state-of-the art in 4 out of 7 metrics.

| Method | Error metrics ↓ | | |
|---|---|---|---|
| | sq. rel | rel | rms |
| Karsch et al. [20]* | 4.894 | 0.417 | 8.172 |
| Laina et al. [25]* | 1.665 | 0.198 | 5.461 |
| **Ours** AdaDepth-S* | 5.71 | 0.452 | 9.559 |
| Godard et al. [13] | 11.990 | 0.535 | 11.513 |
| **Ours** AdaDepth-U | 12.341 | 0.647 | 11.567 |

Table 5. Results on Make3D Dataset. We follow the evaluation scheme used by [13] and compute errors only for depths less than 70 meters. * denotes ground truth supervision.

**Comparison with existing approaches** Interestingly, our unsupervised model *AdaDepth-U* is able to deliver comparable results against previous state-of-the-arts for both NYUD and KITTI natural scenes as shown in Table 3 and Table 4 respectively. Additionally, *AdaDepth-S* outperforms all the previous fully-supervised depth-prediction methods. For a fair comparison with previous adaptation techniques, we also formulated ADDA [45] and DAN [31] (MK-MMD on vectorised convolutional feature) setups for depth adaptation. Training ADDA was very unstable without regularization and we could not get it to converge. Results with DAN are shown in Table 3.

**Generalization to Make3D** To evaluate generalizability of our proposed adaptation method, we adapt the base model trained on Virtual KITTI dataset for the natural scenes of Make3D [38, 39] in both unsupervised and semi-supervised fashion. Apart from the apparent *domain shift* from synthetic to real, there are semantic and physical differences of objects between Virtual KITTI and Make3D datasets. Table 5 shows a quantitative comparison of both *AdaDepth-U* and *AdaDepth-S* on Make3D test set. Evidently, our models generalize and perform reasonably well against the previous arts.

## 6. Conclusion

We present a novel unsupervised domain adaptation method *AdaDepth*, for adapting depth predictions from synthetic RGB-D pairs to natural scenes. We demonstrate *AdaDepth's* efficiency in adapting learned representations from synthetic to real scenes through empirical evaluation on challenging datasets. With the proposed Feature Consistency Framework, *AdaDepth* delivers impressive adaptation results by maintaining spatial content information intact during adversarial learning. While the benchmark results illustrate *AdaDepth's* effectiveness, they also pave way for exploring adjacent problem paradigms. Could a multi-task setting aid in generating richer latent representations for *AdaDepth*? Could we exploit global geometric cues from synthetic images in an efficient way to complement our unsupervised approach? We would like to answer these questions in our future work.

**Acknowledgements** This work was supported by a CSIR Fellowship (Jogendra), and Defence Research and Development Organisation (DRDO), Government of India (Scheme Code: DRDO0672). We also thank Google India for the travel grant.